\documentclass[10pt, conference, compsocconf]{IEEEtran}
\usepackage{multirow}
\usepackage{amsmath}
\usepackage{rotating}
\usepackage[table,usenames,dvipsnames]{xcolor}
\usepackage{graphicx,epsfig}
\usepackage{cite}
\usepackage{url}
\usepackage{color}
\usepackage{paralist}
\usepackage{booktabs}

\begin{document}

\title{A Simplified Description of Fuzzy TOPSIS}

\author{\IEEEauthorblockN{Balwinder Sodhi and Prabhakar T.V.}
\IEEEauthorblockA{Dept. of Computer Science and Engineering,
IIT Kanpur, UP 208016 India\\
\{sodhi, tvp\}@cse.iitk.ac.in}
}

\markboth{Journal of \LaTeX\ Class Files,~Vol.~6, No.~1, January~2012}%
{Shell \MakeLowercase{\textit{et al.}}: Bare Demo of IEEEtran.cls for Computer Society Journals}

\IEEEcompsoctitleabstractindextext{%

\begin{abstract}
A simplified description of Fuzzy TOPSIS (Technique for Order Preference by Similarity to Ideal Situation) is presented. We have adapted the TOPSIS description from existing Fuzzy theory literature and distilled the bare minimum concepts required for understanding and applying TOPSIS. An example has been worked out to illustrate the application of TOPSIS for a multi-criteria group decision making scenario.
\end{abstract}

}

\maketitle

\IEEEdisplaynotcompsoctitleabstractindextext
\IEEEpeerreviewmaketitle

\section{Introduction}
Multi-criteria group decision making (MCGDM) problems are frequently encountered in practice. Several methods exist that can be applied to solve such problems. One example scenario of a MCGDM problem is where a group of two persons (A and B) intends to determine which laptop to buy based on certain criteria. Let's say they have evaluation criteria such as: price, screen size, memory and battery life. Each decision maker can have different importance for different criteria. For example, relative importance of criteria for A can be: price $>$ battery life $>$ memory $>$ screen size. Here $>$ means greater than. For B it can be some different ordering of criteria. Given this scenario, and a set of laptop alternatives, one wants to find which alternative best meets the group's criteria.

Fuzzy TOPSIS is a method that can help in objective and systematic evaluation of alternatives on multiple criteria. In Section \ref{sec:fuzzy_theory_bg} we provide a simplified description of required Fuzzy theory concepts. TOPSIS steps are explained in Section \ref{sec:fuzzy_topsis}, and an example scenario has been worked out in Section \ref{sec:num_illus} to illustrate the TOPSIS steps.

\section{Fuzzy Theory Details}
\label{sec:fuzzy_theory_bg}
A detailed introduction and treatment of the fuzzy set theory is available in \cite{saghafian2005multi,buckley1985ranking}. The definitions of fuzzy concepts that are relevant for understanding of TOPSIS have been adapted from these sources. These definitions are presented as follows.

\paragraph*{Definition 1} A fuzzy set $\tilde{a}$ in a universe of discourse $X$ is characterized by a membership function $\mu_{\tilde{a}}(x)$ that maps each element $x$ in $X$ to a real number in the interval [0, 1]. The function value $\mu_{\tilde{a}}(x)$ is termed the grade of membership of $x$ in $\tilde{a}$. Nearer the value of $\mu_{\tilde{a}}(x)$ to unity, higher will be the grade of membership of $x$ in $\tilde{a}$.

\paragraph*{Definition 2} A triangular fuzzy number is represented as a triplet $\tilde{a} = (a, b, c)$. The membership function $\mu_{\tilde{a}}(x)$ of triangular fuzzy number $\tilde{a}$ is given as:
\begin{equation}
\label{eqn:fuzzy_number_def}
  \mu_{\tilde{a}}(x) = \left\{ 
  \begin{array}{l l}
    \frac{x - a}{b - a} & \quad \text{if } a \leq x \leq b \\
    \frac{c - x}{c - b} & \quad \text{if } b \leq x \leq c \\
    0 & \quad Otherwise \\
  \end{array} \right.
\end{equation}

where $a, b, c$ are real numbers and $a < b < c$. The value of $x$ at $b$ gives the maximal grade of $\mu_{\tilde{a}}(x)$, i.e., $\mu_{\tilde{a}}(x) = 1$; it is the most probable value of the evaluation data. The value of $x$ at $a$ gives the minimal grade of $\mu_{\tilde{a}}(x)$, i.e., $\mu_{\tilde{a}}(x) = 0$; it is the least probable value of the evaluation data. Constants $a$ and $c$ are the lower and upper bounds of the available area
for the evaluation data. These constants reflect the fuzziness of the evaluation data. The narrower the interval $[a, c]$ the lower the fuzziness of the evaluation data.

\subsection{The distance between fuzzy triangular numbers}
Let $\tilde{a} = (a, b, c)$ and $\tilde{b} = (a^\prime, b^\prime, c^\prime)$ be two triangular fuzzy numbers. The distance between them is given using the vertex method by:

\begin{equation}
\label{eqn:fuzzy_distance}
d(\tilde{a}, \tilde{b}) = \sqrt{\frac{1}{3}[(a - a^\prime)^2 + (b - b^\prime)^2 + (c - c^\prime)^2]}
\end{equation}

\subsection{Linguistic variables}
\label{sec:lingu_var_defn}
In fuzzy set theory, conversion scales are applied to transform the linguistic terms into fuzzy numbers. In this paper, we will apply a scale of 1 to 9 for rating the criteria and the alternatives. The linguistic variables and fuzzy ratings for the alternatives and the criteria are as shown in Table-\ref{tab:ling_var_and_fuzzy_rt}.

\begin{table}
\centering
\caption{Fuzzy ratings for linguistic variables}
\label{tab:ling_var_and_fuzzy_rt}
\begin{tabular}{l l l}
\toprule
Fuzzy number & Alternative Assessment & QA Weights\\
\midrule
(1,1,3) & Very Poor (VP) & Very Low (VL) \\
(1,3,5) & Poor (P) & Low (L) \\
(3,5,7) & Fair (F) & Medium (M) \\
(5,7,9) & Good (G) & High (H) \\
(7,9,9) & Very Good (VG) & Very High (VH) \\
\bottomrule
\end{tabular}
\end{table}

The values for the triangular fuzzy numbers that we have chosen for the linguistic variables take into consideration the fuzziness and the distance among the variables (please see equations \ref{eqn:fuzzy_number_def} and \ref{eqn:fuzzy_distance}). The intervals are chosen so as to have a uniform representation from 1 to 9 for the fuzzy triangular numbers used for the five linguistic ratings. For instance, one can also choose (4,5,6) instead of (1,1,3) to represent \textit{Very Low} if one wishes so, however in that case, the ``1 to 9'' ratings would begin from 4 instead of 1. Normalization step takes care of such shifting of the rating scale. The common practice in literature is to start the ratings scales from 1.

\section{Fuzzy TOPSIS}
\label{sec:fuzzy_topsis}
The technique called fuzzy TOPSIS (Technique for Order Preference by Similarity to Ideal Situation) can be used to evaluate multiple alternatives against the selected criteria. In the TOPSIS approach an alternative that is nearest to the Fuzzy Positive Ideal Solution (FPIS) and farthest from the Fuzzy Negative Ideal Solution (FNIS) is chosen as optimal. An FPIS is composed of the best performance values for each alternative whereas the FNIS consists of the worst performance values. A detailed description and treatment of TOPSIS  is discussed by \cite{saghafian2005multi,jiang2008method} and we have adapted the relevant steps of fuzzy TOPSIS as presented below.

Let's say the decision group has $K$ members. If the fuzzy rating and importance weight of the $k$th decision maker, about the $i$th alternative on $j$th criterion, are:\\
$\tilde{x}_{ij}^k = (a_{ij}^k, b_{ij}^k, c_{ij}^k)$ 
and $\tilde{w}_j^k = (a^{\prime k}_j, b^{\prime k}_j, c^{\prime k}_j)$ respectively, \\
where $i = 1,2,\dots, m,$ and $j = 1,2,\dots, n$, 
then the aggregated fuzzy ratings $\tilde{x}_{ij}$ of alternatives ($i$) with respect to each criterion ($j$) are given by $\tilde{x}_{ij} = (a_{ij}, b_{ij}, c_{ij})$ such that:

\begin{equation}
\label{eqn:topsis_aggregate_ratings}
a_{ij} = \min_{k}\{a_{ij}^k\}, \quad b_{ij} = \frac{1}{K} \sum_{k=1}^K b_{ij}^k, \quad c_{ij} = \max_{k}\{c_{ij}^k\}
\end{equation}

The aggregated fuzzy weights of each criterion are calculated as $\tilde{w}_j = (a^\prime_j, b^\prime_j, c^\prime_j)$ where:
\begin{equation}
\label{eqn:topsis_aggregate_wt}
a^\prime_j = \min_{k}\{a_j^{\prime k}\}, \quad b^\prime_j = \frac{1}{K} \sum_{k=1}^K b_j^{\prime k}, \quad c^\prime_j = \max_{k}\{c_j^{\prime k}\}
\end{equation}
\\

A fuzzy multicriteria Group Decision Making (GDM) problem which can be concisely expressed in matrix format as:
\begin{equation}
\label{eqn:topsis_dm1}
\tilde{D} = \bordermatrix{~ & C_1 & C_2 & & C_n \cr
                  A_1 & \tilde{x}_{11} & \tilde{x}_{12} & \dots & \tilde{x}_{1n} \cr
                  A_2 & \tilde{x}_{21} & \tilde{x}_{22} & \dots & \tilde{x}_{2n} \cr
                      & \dots & \dots & \tilde{x}_{ij} & \dots \cr
                  A_m & \tilde{x}_{m1} & \tilde{x}_{m2} & \dots & \tilde{x}_{mn} \cr}
\end{equation}
\begin{equation}
\label{eqn:topsis_dm2}
\tilde{W} = (\tilde{w}_1, \tilde{w}_2,\dots, \tilde{w}_n)
\end{equation}
  
where for all $\tilde{x}_{ij} $ and $\tilde{w}_{j}, i=1,2,\dots, m; j=1,2,\dots, n$. Here $\tilde{x}_{ij} = (a_{ij}, b_{ij}, c_{ij})$ and $\tilde{w}_j = (a^\prime_j, b^\prime_j, c^\prime_j)$ are triangular fuzzy numbers representing linguistic variables. To keep the normalization formula simple, the linear scale transformation is used to transform various criteria scales into a comparable scale. Thus, we have the normalized fuzzy decision matrix as:

\begin{equation}
\label{eqn:topsis_dm_normalized1}
\tilde{R} = [\tilde{r}_{ij}]_{m \times n}, i=1,2,\dots, m; j=1,2,\dots, n
\end{equation}
where:

\begin{equation}
\label{eqn:topsis_dm_normalized2}
\left.
\begin{aligned}
\tilde{r}_{ij} &= \left(\frac{a_{ij}}{c_j^*}, \frac{b_{ij}}{c_j^*}, \frac{c_{ij}}{c_j^*}\right) \quad \text{and}\\
c_j^* &= \max_i c_{ij} \quad \text{(benefit criteria)}
\end{aligned}
\quad \right\}
\end{equation}

\begin{equation}
\label{eqn:topsis_dm_normalized3}
\left.
\begin{aligned}
\tilde{r}_{ij} &= \left(\frac{a_j^-}{c_{ij}}, \frac{a_j^-}{b_{ij}}, \frac{a_j^-}{a_{ij}}\right) \quad \text{and}\\
a_j^- &= \min_i a_{ij} \quad \text{(cost criteria)}
\end{aligned}
\quad \right\}
\end{equation}
\\

The above normalization method preserves the property that the ranges of normalized triangular fuzzy numbers belong to $[0, 1]$. 

The weighted normalized fuzzy decision matrix $\tilde{V}$ is computed by multiplying the weights ($\tilde{w}_j$) of evaluation criteria with the normalized fuzzy decision matrix $\tilde{r}_{ij}$ as:
\begin{equation}
\label{eqn:topsis_dm_weighted}
\left.
\begin{aligned}
\tilde{V} = & [\tilde{v}_{ij}]_{m\times n}, \text{where:} \\
i = & 1, 2,\dots, m; j = 1, 2,\dots, n. \\
\tilde{v}_{ij} = & \tilde{r}_{ij}(\cdot)\tilde{w}_j = (a^{\prime\prime}_{ij}, b^{\prime\prime}_{ij}, c^{\prime\prime}_{ij})
\end{aligned}
\right\}
\end{equation}
\\

The FPIS and FNIS of the alternatives are defined as follows:
\begin{equation}
\label{eqn:topsis_fpis}
\left.
\begin{aligned}
A^* = & (\tilde{v}_1^*, \tilde{v}_2^*, \dots, \tilde{v}_n^*) \quad \text{where:}\\
\tilde{v}_j^*= & (c, c, c) \text{ such that: }\\
c = &\max_i\{c^{\prime\prime}_{ij}\}, i = 1, 2\dots, m; j = 1, 2,\dots, n
\end{aligned}
\quad \right\}
\end{equation}

\begin{equation}
\label{eqn:topsis_fnis}
\left.
\begin{aligned}
A^- = & (\tilde{v}_1^-, \tilde{v}_2^-, \dots, \tilde{v}_n^-) \quad \text{where:} \\
\tilde{v}_j^- = & (a, a, a) \text{ such that: } \\
a = & \min_i\{a^{\prime\prime}_{ij}\}, i = 1, 2\dots, m; j = 1, 2,\dots, n \\
\end{aligned}
\quad \right\}
\end{equation}
\\

The distance ($d_i^*$ and $d_i^-$) of each weighted alternative $i = 1, 2,\dots, m$ from the FPIS and the FNIS is computed as follows:
\begin{equation}
\label{eqn:topsis_fpis_dist}
d_i^* = \sum_{j=1}^n d_v(\tilde{v}_{ij}, \tilde{v}_j^*),\quad i = 1, 2,\dots, m
\end{equation}

\begin{equation}
\label{eqn:topsis_fnis_dist}
d_i^- = \sum_{j=1}^n d_v(\tilde{v}_{ij}, \tilde{v}_j^-),\quad i = 1, 2,\dots, m
\end{equation}
where $d_v(\tilde{a}, \tilde{b})$ is the distance measurement between two fuzzy numbers $\tilde{a}$ and $\tilde{b}$.
\\

The closeness coefficient $CC_i$ represents the distances to fuzzy positive ideal solution, $A^*$, and the fuzzy negative ideal solution, $A^-$ simultaneously. The closeness coefficient of each alternative is calculated as:
\begin{equation}
\label{eqn:topsis_cc}
CC_i = \frac{d_i^-}{d_i^- + d_i^*},\quad i = 1, 2\dots, m
\end{equation}
\\

The alternative with highest closeness coefficient represents the best alternative and is closest to the FPIS and farthest from the FNIS. In summary, the TOPSIS steps that we use are as follows:

\begin{enumerate}[i.]
	\item Aggregate the weight of criteria to get the aggregated fuzzy weight $\tilde{w}_j$ of criterion $C_j$ (using equation \ref{eqn:topsis_aggregate_wt}) and pool the decision makers' ratings to get the aggregated fuzzy rating $\tilde{x}_{ij}$ of alternative $A_i$ under criterion $C_j$ (using equation \ref{eqn:topsis_aggregate_ratings}).
	\item Construct the fuzzy decision matrix and the normalized fuzzy decision matrix (equations \ref{eqn:topsis_dm_normalized1}, \ref{eqn:topsis_dm_normalized2} and \ref{eqn:topsis_dm_normalized3}).
	\item Construct the weighted normalized fuzzy decision matrix (equation \ref{eqn:topsis_dm_weighted}).
	\item Determine FPIS and FNIS and calculate the distance of each alternative from FPIS and FNIS, respectively (equations \ref{eqn:topsis_fpis}, \ref{eqn:topsis_fnis}, \ref{eqn:topsis_fpis_dist} and \ref{eqn:topsis_fnis_dist}).
	\item Calculate the closeness coefficient of each alternative and rank the alternatives (equation \ref{eqn:topsis_cc}).
\end{enumerate}

\subsection{TOPSIS calculator}
We have developed a program in Java which implements the steps described in this paper. Input to this tool is a CSV file (which is easy to edit in a spreadsheet program such as LibreOffice or Excel) where a user can specify complete set of linguistic inputs etc. The program is available at \texttt{https://bitbucket.org/sodhi/topsistool} as open source software.

\section{Numerical Illustration}
\label{sec:num_illus}
An illustrative application of TOPSIS steps, as discussed in preceding sections, for a scenario involving 3 decision makers, 4 evaluation criteria C1 -- C4 (all of benefit type), and rating scale is as shown in Table \ref{tab:ling_var_and_fuzzy_rt} is described below.

In this example, a team of three decision makers $D_1, D_2$ and $D_3$ is formed to evaluate the two alternatives, $A_1$ and $A_2$, for picking the optimal one. Key input from decision makers is typically to identify the proper weightage to various criteria.

The team provided linguistic weightage for the criteria in Table \ref{tab:uc_criteria_wt}, and assessment for three alternatives on each of the criteria item is presented in Table \ref{tab:uc_alt_ratings}.

Results of various TOPSIS calculation steps are shown in Tables \ref{tab:aggr_fuzzy_dm} -- \ref{tab:uc_distances}. Closeness coefficients, $CC_i $, of the two alternatives $A_1$ and $A_2$ come out to be 0.477 and 0.454 respectively. Hence the ranking order for the alternatives is $A_1 > A_2$, that is, $A_1$ is the best choice considering the given criteria. The closeness coefficient scores for alternatives are numeric values and can be further utilized to indicate the degree of inferiority or superiority of the alternatives w.r.t each other.

\begin{table}
	\setlength{\tabcolsep}{4 pt}
	\centering
	\caption{Criteria Weightage By Decision Makers\label{tab:uc_criteria_wt}}
	\begin{tabular}{l*{4}{c}}
		\toprule
		& C1 & C2 & C3 & C4 \\ \midrule
		$DM_1$ & H (5, 7, 9) & VH (7, 9, 9) & VH (7, 9, 9) & M (3, 5, 7)\\ 
		$DM_2$ & M (3, 5, 7) & H (5, 7, 9) & H (5, 7, 9) & L (1, 3, 5)\\ 
		$DM_3$ & M (3, 5, 7) & H (5, 7, 9) & H  (5, 7, 9)& L (1, 3, 5)\\ \bottomrule
	\end{tabular}
\end{table}

\begin{table}
	\centering
	\caption{Alternatives Ratings By Decision Makers\label{tab:uc_alt_ratings}}
	\begin{tabular}{l*{6}{c}}
		\toprule
		\textbf{Criteria} & \multicolumn{3}{c}{\textbf{A1}} & \multicolumn{3}{c}{\textbf{A2}} \\
		& DM1 & DM2 & DM3 & DM1 & DM2 & DM3  \\ \midrule
		C1 & F & F & F & G & G & F \\
		C2 & VG & VG & VG & G & VG & G\\ 
		C3 & P & F & P & P & P & P\\
		C4 & F & F & P & P & P & F\\ \bottomrule
	\end{tabular}
\end{table}

\begin{table*}
	\setlength{\tabcolsep}{4 pt}
	\centering
	\caption{Aggregate Fuzzy Decision Matrix \label{tab:aggr_fuzzy_dm}}
	\begin{tabular}{l*{4}{c}}
		\toprule
		Alternatives & C1 & C2 & C3 & C4 \\ \midrule
		
		A1 & (3.000, 5.000, 7.000) & (7.000, 9.000, 9.000) & (1.000, 3.667, 7.000) & (1.000, 4.333, 7.000) \\
		A2 & (3.000, 6.333, 9.000) & (5.000, 7.667, 9.000) & (1.000, 3.000, 5.000) & (1.000, 3.667, 7.000) \\
		\bottomrule
	\end{tabular}
\end{table*}

\begin{table*}
	\setlength{\tabcolsep}{4 pt}
	\centering
	\caption{Normalized aggregate Fuzzy Decision Matrix \label{tab:aggr_norm_fuzzy_dm}}
	\begin{tabular}{l*{4}{c}}
		\toprule
		Alternatives & C1 & C2 & C3 & C4 \\ \midrule
		
		A1 & (0.333, 0.556, 0.778) & (0.778, 1.000, 1.000) & (0.143, 0.524, 1.000) & (0.143, 0.619, 1.000)\\
		A2 & (0.333, 0.704, 1.000) & (0.556, 0.852, 1.000) & (0.143, 0.429, 0.714) & (0.143, 0.524, 1.000)\\
		\bottomrule
	\end{tabular}
\end{table*}

\begin{table*}
	\centering
	\caption{Weighted Normalized Fuzzy Decision Matrix\label{tab:uc_wtd_norm_fuzzy_dm}}
	\begin{tabular}{l*{4}{c}}
		\toprule
		Alternatives & C1 & C2 & C3 & C4 \\ \midrule
		
		A1 & (1.000, 3.148, 7.000) & (3.889, 7.667, 9.000) & (0.714, 4.016, 9.000) & (0.143, 2.270, 7.000) \\
		A1 & (1.000, 3.988, 9.000) & (2.778, 6.531, 9.000) & (0.714, 3.286, 6.429) & (0.143, 1.921, 7.000) \\
		\bottomrule
	\end{tabular}
\end{table*}

\begin{table}
	\setlength{\tabcolsep}{4 pt}
	\centering
	\caption{Distances $d_v(A_i, A^*)$ and $d_v(A_i, A^-)$ from FPIS and FNIS for alternatives \label{tab:uc_distances}}
	
	\begin{tabular}{l *{4}{c}}
		\toprule
		\textbf{Criteria} & \textbf{FPIS($A_1$)}  & \textbf{FPIS($A_2$)}  &  \textbf{FNIS($A_1$)}  & \textbf{FNIS($A_2$)} \\ \midrule
		C1 & 5.837 & 5.45 & 3.679 & 4.93	\\ 
		C2 & 3.049 & 3.864 &	4.613 & 4.195	\\
		C3 & 5.582 & 5.997	& 5.149 & 3.617	\\ 
		C4 & 4.809 & 4.926 & 4.145 & 4.089	\\ \bottomrule
	\end{tabular}
\end{table}

\section*{Acknowledgments}
We would like to thank Dr. Alessandro de Castro Correa of Federal Institute of Education, Science and Technology of Para (IFPA), Brazil for highlighting possible improvements to the example discussed in previous version of this paper. We also thank Ritu Kapur of IIT Ropar for helping with the example data verification and pointing out some typos.

\end{document}